%% file: Manuscript.tex
\documentclass{article}
\usepackage{spconf,amsmath,graphicx}
\usepackage{multirow}
\usepackage{booktabs}
\usepackage{tabularray}
\usepackage{subcaption}
\usepackage[shortlabels]{enumitem}
\usepackage[hyphens]{url}
\usepackage[hidelinks]{hyperref}
\usepackage{cite}

\title{OpenRR-1k: A Scalable Dataset for Real-World Reflection Removal}
\name{Kangning Yang, Ling Ouyang, Huiming Sun, Jie Cai\textsuperscript{*}\thanks{* Project Lead}, Lan Fu, Jiaming Ding, Chiu Man Ho, Zibo Meng}
\address{OPPO AI Center}

\begin{document}
%
\maketitle
\begin{abstract}
Reflection removal technology plays a crucial role in photography and computer vision applications. However, existing techniques are hindered by the lack of high-quality in-the-wild datasets. In this paper, we propose a novel paradigm for collecting reflection datasets from a fresh perspective. Our approach is convenient, cost-effective, and scalable, while ensuring that the collected data pairs are of high quality, perfectly aligned, and represent natural and diverse scenarios. Following this paradigm, we collect a Real-world, Diverse, and Pixel-aligned dataset (named OpenRR-1k dataset), which contains 1,000 high-quality transmission-reflection image pairs collected in the wild. Through the analysis of several reflection removal methods and benchmark evaluation experiments on our dataset, we demonstrate its effectiveness in improving robustness in challenging real-world environments. Our dataset is available at \url{https://github.com/caijie0620/OpenRR-1k}.

\end{abstract}

\input{1-Introduction}
\input{2-RelatedWork}
\input{3-Methods}
\input{4-Experiments}
\input{5-Conclusion}

\bibliographystyle{IEEEbib}
\bibliography{refs}

\end{document}

%% file: 1-Introduction.tex
\section{Introduction}
\label{sec:intro}
Single image reflection removal (SIRR) is a critical task in image processing, focusing on recovering the true scene behind reflections from reflective surfaces (e.g., transparent glasses). This is crucial in fields such as autonomous driving, medical imaging~\cite{yue2021automatic, shih2023deep}, and augmented reality. Over the years, various techniques have been proposed, ranging from traditional image decomposition methods to more advanced deep learning-based approaches~\cite{fan2017generic, li2020single, song2023robust}.

\begin{figure}[!t]
    \centering
    \footnotesize
    \includegraphics[width=0.35\textwidth]{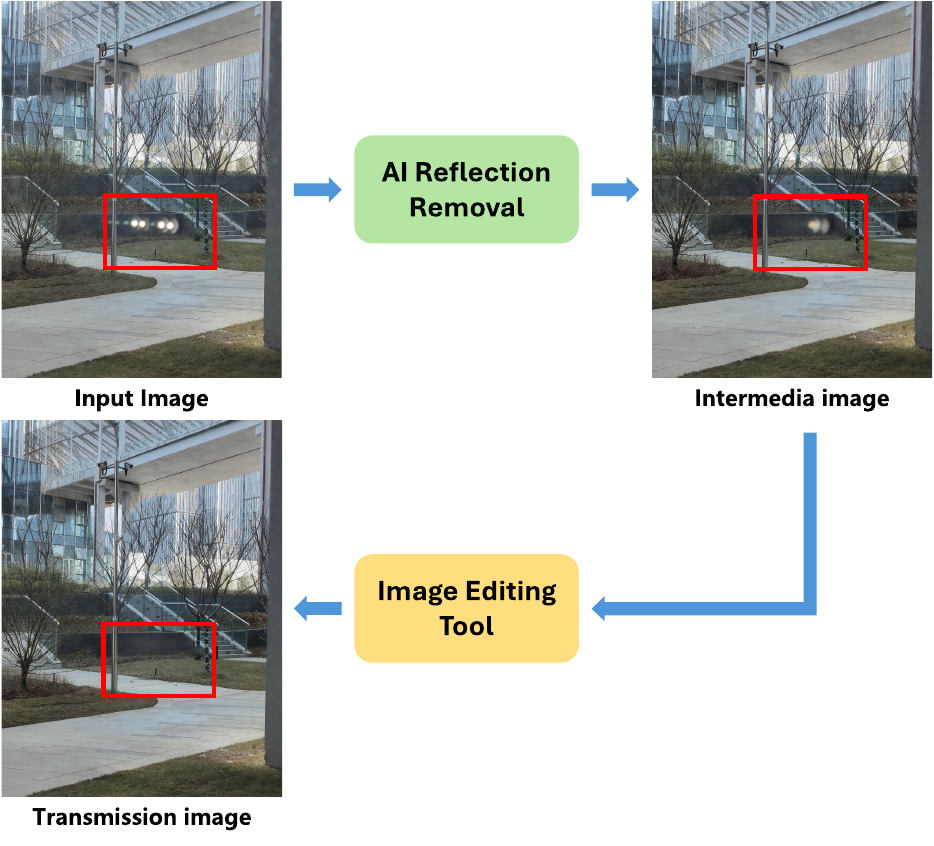}
    \caption{Visualization of paired data generation pipeline.}
    \vspace{-20pt}
    \label{fig:protocol}
\end{figure}

\begin{figure*}[h]
    \centering
    \footnotesize
    \includegraphics[width=0.78\textwidth]{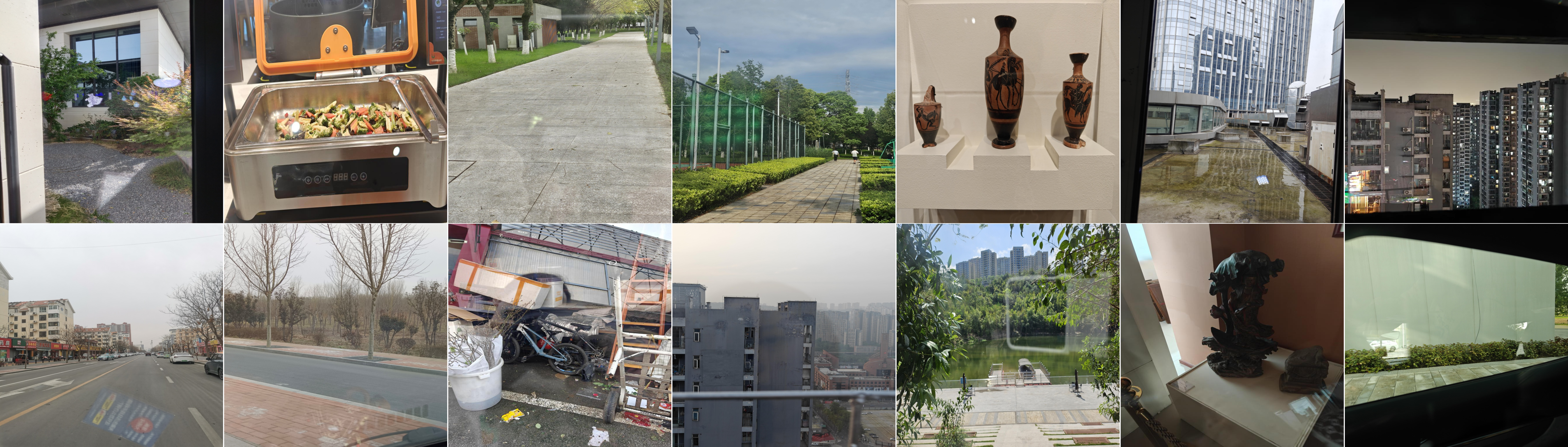}
    \caption{Overview of our OpenRR-1k dataset. }
    \vspace{-15pt}
    \label{fig:dataset_examples} 
\end{figure*}

Despite significant efforts and progress made in modeling the formation of reflection contamination, the lack of high-quality data has increasingly become a bottleneck, limiting the full potential of deep learning models. Given that a sufficient supply of high-quality data is crucial for the success of data-driven approaches, we propose a novel data collection protocol specifically designed to capture high-quality pairs of transmission and blended images. Specifically, to ensure a diverse and natural set of real-world images, we impose no restrictions on image capture. Researchers are free to recruit individuals to capture images featuring various types of reflection lights, irrespective of their angles, shapes, intensities, materials, or other characteristics. Alternatively, images with reflections can be sourced directly from social media platforms such as Flickr, Instagram, and others. Building on this, to ensure strict alignment between the ground-truth transmission images and the blended images with reflection contamination, we rely on proven and effective tools for reflection removal, instead of directly or indirectly obtaining ground-truth transmission images by removing the glass or using light-absorbing black cloth as done in previous methods~\cite{li2020single, wan2017benchmarking, zhang2018single, lei2020polarized, lei2022categorized, zhu2024revisiting}. To guarantee the accuracy of the ground truth, we further propose using image editing tools to eliminate any remaining reflection residues. While this step still involves some manual post-processing, it is considerably more practical, cost-effective, and convenient compared to existing methods. Furthermore, this approach can leverage crowdsourcing platforms like Amazon Mechanical Turk to streamline the process and enhance scalability. Following this protocol, we developed a new high-quality dataset, the OpenRR-1k dataset, aimed at advancing research in reflection removal. To evaluate the effectiveness of the OpenRR-1k dataset, we conduct a comprehensive benchmark evaluation and further analyze the limitations of existing methods. The contributions of our work are three-fold: 
\begin{itemize}[noitemsep, nolistsep,leftmargin=*] 
    \item We proposed a new data collection protocol to collect diverse and well-aligned real-world reflection-transmission image pairs. The proposed protocol is highly scalable, enabling the creation of larger-scale datasets in the future. 
    \item We collected the OpenRR-1k dataset, a high-quality collection consisting of 1,000 in-the-wild image pairs. 
    \item We conducted a comprehensive benchmark evaluation within the SIRR domain. Additionally, we demonstrated the improvements enabled by our OpenRR-1k dataset when applied to existing reflection removal approaches. 
\end{itemize}

%% file: 2-RelatedWork.tex
\section{Related Work}
\label{sec:rel}
Currently, public datasets supporting SIRR can be divided into two categories: fully-synthetic and semi-synthetic datasets. For fully-synthetic datasets, this technique typically involves selecting two clean images (without reflections) and combining them with different coefficients to create a synthetic image with reflections, which serves as the corresponding blended image. This process is then used to generate a large number of paired blended-transmission data samples~\cite{fan2017generic, zhang2018single, zhang2023benchmarking}. For example, Guo et al.\cite{guo2014robust} simply set 0.6 for transmission and 0.4 for reflection images. Fan et al.\cite{fan2017generic} assumed that the reflection is relatively blurry compared to the sharper and clearer background layer, and synthesized images by combining the background and reflection layers while avoiding brightness overflow through adaptive subtraction and clipping instead of scaling. They also adjusted the Gaussian blur kernel $\sigma$ to handle a wider range of reflection blurriness, including cases with less blurry reflections. Zhang et al.\cite{zhang2023benchmarking} focused on ultra-high-definition image synthesis. 

For semi-synthetic datasets, researchers typically rely on props such as glass and cloth. After capturing the blended images, they create reflection pairs by either removing the glass or blocking the background or reflection light with light-absorbing black velvet cloth~\cite{li2020single, wan2017benchmarking, zhang2018single, lei2020polarized, lei2022categorized, zhu2024revisiting}. For example, Li et al.\cite{li2020single} obtained transmitted images by manually removing the glass, while Lei et al.\cite{lei2020polarized, lei2022categorized} focused on the RAW data space and indirectly obtained transmission images through subtraction. More recently, Zhu et al.~\cite{zhu2024revisiting} proposed a novel data collection pipeline that involves blocking all reflection lights generated by the surrounding environment.

Nevertheless, synthetic-based methods, while useful, come with inherent limitations. Fully-synthetic approaches are typically developed based on a range of assumptions about the scene and the underlying physical processes. As a result, their performance is often constrained by the domain gap between synthetic and real-world data. The assumptions made during the creation of synthetic data are often overly simplified, which can lead to suboptimal performance when these methods are applied to real-world images~\cite{lei2022categorized}. On the other hand, manual methods for glass removal, as discussed in~\cite{zhu2024revisiting}, face challenges such as spatial pixel misalignment. This misalignment arises from refraction caused by the glass, which distorts the captured image. The subtraction method in the RAW data space, which aims to eliminate reflections, may also fail to completely remove the reflective areas, leaving residual artifacts in the transmission images. Even the methods proposed in~\cite{zhu2024revisiting}, though promising, encounter difficulties when deployed in real-world environments. One significant challenge is the uncontrollable nature of environmental factors~\cite{ntire2025reflection, yang2025survey}, such as wind or equipment vibrations, which can lead to misalignment between the transmission image $\mathbf{T}$ and the blended image $\mathbf{I}$. Additionally, while black cloth is used to block local object light and global ambient light, achieving a perfect blockade is practically difficult, resulting in color discrepancies between $\mathbf{T}$ and $\mathbf{I}$. Another limitation of current synthetic-based methods is that they rely heavily on pre-defined reflection scenes and content. This not only makes the data collection process time-consuming and labor-intensive but also restricts the diversity of the collected data. Although~\cite{zhu2024revisiting} proposed dynamic manipulation of reflective content, these reflections are still not entirely natural, as they do not spontaneously occur in real-world settings. Furthermore, replicating intensity, shape, and color of reflections in a manner that faithfully represents the scene's geometry and lighting conditions remains a challenge.

%% file: 3-Methods.tex
\section{Methodology}
\label{sec:method}
\subsection{Dataset Collection Protocol}
As shown in Fig.~\ref{fig:protocol}, our proposed data collection protocol consists of two main steps. The first step involves using a proven off-the-shelf tool to initially remove reflections from the images. We adopted the OPPO smartphone's AI-based reflection removal software to obtain the initial reflection removal results. This commercial software, integrated into OPPO smartphones, is specifically designed to handle reflection artifacts in photographs and is one of the few effective tools currently available on the market for this purpose, with similar tools only offered by Samsung\footnote{\href{https://www.samsung.com/latin_en/support/mobile-devices/how-to-use-the-galaxy-s23-object-eraser/}{https://www.samsung.com/latin\_en/support/mobile-devices/}} and Huawei\footnote{\href{https://consumer.huawei.com/en/support/content/en-us15870917/}{https://consumer.huawei.com/en/support/}}.

We observed that the initial reflection removal results removed major reflection components, however, subtle residual reflections remain, as shown in the intermediate image of Fig.~\ref{fig:protocol}. To address this, the second stage of our protocol involves a refinement process to recover more details.  Spefically, we adopted professional image editing tools (e.g., Photoshop, MeituPic, etc.) for the refinement. This step is crucial for eliminating any remaining artifacts or inconsistencies in the intermediate images. After precise manual adjustments, the final processed images are of high quality and suitable for training and evaluation purposes. This manual intervention allows us to preserve fine details while eliminating any potential artifacts introduced during AI processing. To this end, we defined three manual evaluation criteria:
\begin{itemize}[noitemsep, nolistsep,leftmargin=*]
\item \textbf{Cleanliness}: Effectively remove both strong and weak reflections without visible residuals.
\item \textbf{Artifacts}: Avoid unintended content removal or unnatural distortions.
\item \textbf{Overall Image Quality}: The restored image should surpass the input in perceptual quality, including color fidelity, texture, detail preservation, and overall visual coherence.
\end{itemize}

Compared to existing data collection methods~\cite{li2020single, wan2017benchmarking, zhang2018single, lei2020polarized, lei2022categorized, zhu2024revisiting}, our approach offers several key advantages:
1) \textbf{Diversity}: Our method allows for the collection of a significantly broader range of data samples, without being restricted by specific lighting conditions or types of glass surfaces. The collected images can cover various real-world reflection scenarios, including diverse lighting conditions (e.g., daylight, sunset, and nighttime illumination) and different glass surfaces, such as car windows, building glass doors, museum display cases, and other types of glass (see Fig.~\ref{fig:dataset_examples}).  2) \textbf{Pixel-level Alignment}: Using off-the-shelf tools, we ensure that the input images with reflections and the processed transmission images are perfectly aligned. 3) \textbf{True Real-World Data}: Since our method does not require the collection of ground-truth data, we can capture reflection scenarios directly from real-world environments, without relying on artificial setups or simulated reflections. This enables us to collect data that truly represents genuine real-world situations.

\subsection{OpenRR-1k Dataset}
Based on our proposed protocol, we constructed the OpenRR-1k dataset, which consists of a total of 1,000 image pairs. The dataset is partitioned into training, validation, and test sets with an 80/10/10 split, respectively.

\input{Tables/table1}

\input{Tables/table2}

Table~\ref{tab:1} presents a comprehensive comparison between our OpenRR-1k dataset and other publicly available reflection removal datasets. Compared to SIR$^2$\cite{wan2017benchmarking}, Real\cite{zhang2018single}, and Nature~\cite{li2020single}, our OpenRR-1k dataset offers a greater number of image pairs and higher image resolution. While RRW~\cite{zhu2024revisiting} contains more data pairs and images of higher resolution, we argue that our dataset is more scalable and facilitates faster, more convenient collection of high-quality data. This advantage stems from the fact that it does not require specialized data collection equipment or consideration of various environmental factors. In fact, when attempting to use the RRW pipeline, we found it difficult to operate in a real deployment. 

\begin{figure}[h]
    \centering
    \footnotesize
    \includegraphics[width=0.5\textwidth]{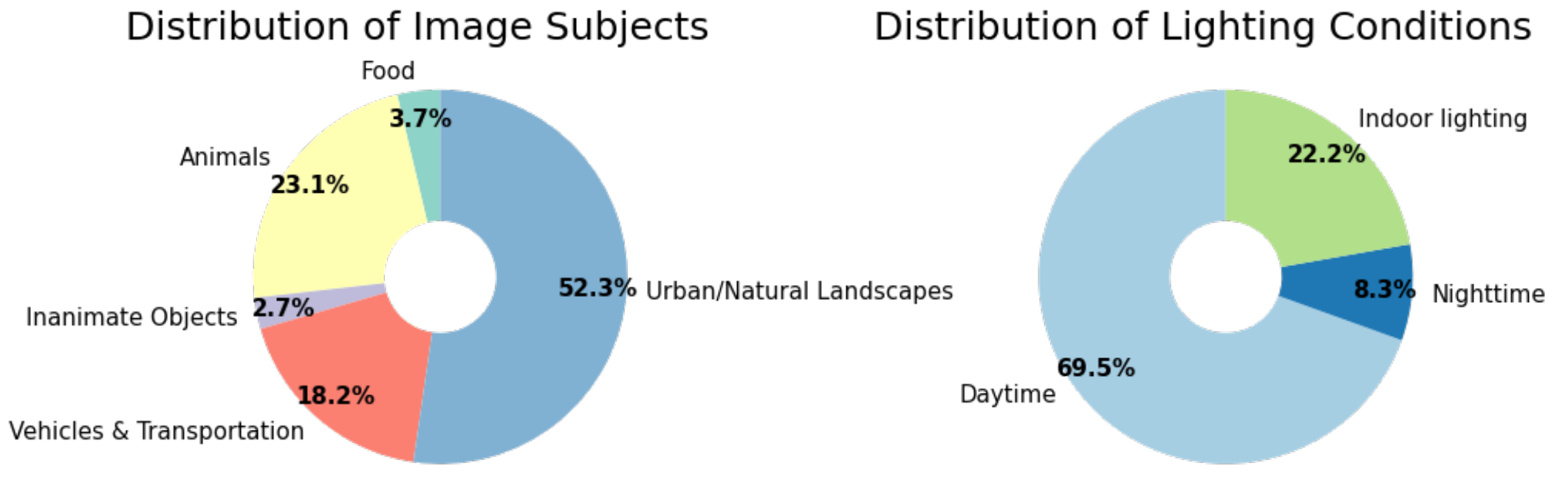}
    \caption{The category distribution of our OpenRR-1k dataset}
    \vspace{-5pt}
    \label{fig:statistics}	 
\end{figure}

Fig.~\ref{fig:statistics} provides an overview of the categorical composition of our OpenRR-1k dataset from two perspectives: scene content and lighting conditions. In terms of scene content (as shown in the left pie chart), we categorized the dataset into five main groups: food, animals, inanimate objects, vehicles \& transportation, and urban/natural landscapes. Regarding lighting distribution (as shown in the right pie chart), we divided it across three distinct scenarios: daytime, nighttime, and indoor lighting. 

%% file: Tables/table1.tex
\begin{table}[ht]
\centering
\footnotesize
\caption{Comparison of existing datasets with our OpenRR-1k dataset}
\resizebox{0.4\textwidth}{!}{
\begin{tabular}{ccccc}
\toprule
Dataset & Year & Usage      & \begin{tabular}[c]{@{}c@{}}Pair \\ Number\end{tabular} & \begin{tabular}[c]{@{}c@{}}Average \\ Resolution\end{tabular} \\ \hline
SIR$^2$~\cite{wan2017benchmarking}   & 2017 & Test       & 454         & 540 x 400          \\
Real~\cite{zhang2018single}        & 2018 & Train/Test & 89/20       & 1152 x 930         \\
Nature~\cite{li2020single}      & 2020 & Train/Test & 200/20      & 598 x 398          \\
RRW~\cite{zhu2024revisiting}      & 2023 & Train & 14952      & 2580 × 1460          \\
\textbf{Ours}        & 2025 & Train/Val/Test & 800/100/100    &  922 x 917                  \\ \bottomrule
\end{tabular}
}
\vspace{-10pt}
\label{tab:1}
\end{table}

%% file: Tables/table2.tex
\begin{table*}[ht]
\centering
\footnotesize
\caption{Quantitative Comparisons of Real-World Reflection Benchmarks. The best results are in \textbf{bold}, and the second-best results are \underline{underlined}.}
\resizebox{1\textwidth}{!}{

\begin{tabular}{ccccccccccccccccc}
\toprule
\multirow{2}{*}{\begin{tabular}[c]{@{}c@{}}Training \\ Strategy\end{tabular}} & \multirow{2}{*}{Methods} & \multicolumn{3}{c}{\textit{Nature} (20)} & \multicolumn{3}{c}{\textit{Real} (20)} & \multicolumn{3}{c}{\textit{SIR\textsuperscript{2}} (454)} & \multicolumn{3}{c}{$\textit{OpenRR\text{-}1k}_{\textit{val}}$ (100)} & \multicolumn{3}{c}{$\textit{OpenRR\text{-}1k}_{\textit{test}}$ (100)} \\ \cmidrule(lr){3-5} \cmidrule(lr){6-8} \cmidrule(lr){9-11} \cmidrule(lr){12-14} \cmidrule(lr){15-17} 
                                   &                                               & PSNR $\uparrow$           & SSIM $\uparrow$          & LPIPS $\downarrow$          & PSNR $\uparrow$         & SSIM $\uparrow$        & LPIPS $\downarrow$          & PSNR $\uparrow$         & SSIM $\uparrow$          & LPIPS $\downarrow$          & PSNR $\uparrow$          & SSIM $\uparrow$          & LPIPS $\downarrow$          & PSNR $\uparrow$          & SSIM $\uparrow$ & LPIPS $\downarrow$          \\ \hline
            -                      & Input Image                    & 20.46          & 0.789          & 0.167          & 19.07         & 0.758        & 0.240          & 22.76         & 0.892         & 0.133          & 26.37               & 0.943               & 0.077          & 26.20                 & 0.939     & 0.078          \\ \midrule
\multirow{5}{*}{S1}                & ERRNet~\cite{wei2019single}                   & 20.56          & 0.783          & 0.170          & 23.10         & \textbf{0.831}        & \underline{0.152}          & 23.36         & 0.893         & 0.124          & 24.08             & 0.928         & 0.098          & 24.14            & 0.919 & 0.100          \\
                                   & DSRNet~\cite{hu2023single}                    & 25.07          & 0.839          & \underline{0.113}          & \textbf{23.85}         & \underline{0.826}        & \textbf{0.146}          & \textbf{25.69}         & \textbf{0.926}         & \textbf{0.088}          & \textbf{25.35}               & \underline{0.929}               & 0.100          & \textbf{25.76}               & \underline{0.923}               & 0.098          \\
                                   & RAGNet~\cite{li2023two}                       & 20.50          & 0.778          & 0.171          & 21.07         & 0.789        & 0.199          & 24.65         & 0.900         & 0.115          & 25.01               & 0.926             & \underline{0.094}          & 25.12 & \underline{0.923} & \underline{0.094}          \\
                                   & RDNet-RRNet~\cite{zhu2024revisiting}  & \textbf{26.04} & \textbf{0.846} & \textbf{0.100}          & 21.83          & 0.801         & 0.177          & \underline{25.49} & \underline{0.916}          & \underline{0.110}          & \underline{25.17}               & \textbf{0.934}             & \textbf{0.086}          & \underline{25.66}	& \textbf{0.935} & \textbf{0.083}          \\ 
                                    & Ours          & \underline{25.44}  & \underline{0.844}        & 0.115          & \underline{23.25}       & 0.824       & 0.163          & 22.93  & 0.882  & 0.156          & 24.61          & 0.918      & 0.108          & 24.15	     &  0.910       & 0.113          \\ \midrule
\multirow{5}{*}{S2}                & ERRNet~\cite{wei2019single}                   & 20.41               & 0.789               & 0.179          & 23.64             & 0.829       & 0.149          & 23.39              & 0.899              & 0.120          & 27.26               & 0.949             & 0.070          & 28.19               &  0.951             & 0.061           \\
                                   & DSRNet~\cite{hu2023single}                    & \underline{26.33}          & \underline{0.853}         & \textbf{0.105}          & \textbf{25.00}         & \textbf{0.844}        & \textbf{0.131}          & \textbf{25.83}        & \textbf{0.928}          & \textbf{0.080}          & 29.11         & \underline{0.959}            & \underline{0.053}          & 29.22         & 0.957            & \underline{0.051}          \\
                                   & RAGNet~\cite{li2023two}                       & 21.11          & 0.789         & 0.164          & 21.17         & 0.788        & 0.189         & 24.49        & 0.903          & 0.111          & 26.94         & 0.943           & 0.073          & 27.11	& 0.941 & 0.071          \\
                                   & RDNet-RRNet~\cite{zhu2024revisiting}  & \textbf{26.34}          & \textbf{0.857}         & \underline{0.112}          & 23.16         & 0.811        & 0.167          & 24.57        & 0.911          & 0.114          & \underline{29.24}         & 0.957           & 0.060          & \underline{30.27}	& \underline{0.960} & 0.053          \\
                                   & Ours                                          &  26.09         & 0.851          & \textbf{0.105}          & \underline{24.29}         & \underline{0.834}        & \underline{0.139}          & \underline{24.71}        & \underline{0.915}         & \underline{0.106}          & \textbf{31.74}         & \textbf{0.965}           & \textbf{0.040}          & \textbf{31.93}	& \textbf{0.964} & \textbf{0.038}          \\ \bottomrule
\end{tabular}
}
\vspace{-10pt}
\label{tab:2}
\end{table*}

%% file: 4-Experiments.tex
\section{Experiments}
\label{sec:expe}

\begin{figure*}[ht]
    \centering
    \footnotesize
    \includegraphics[width=0.8\textwidth]{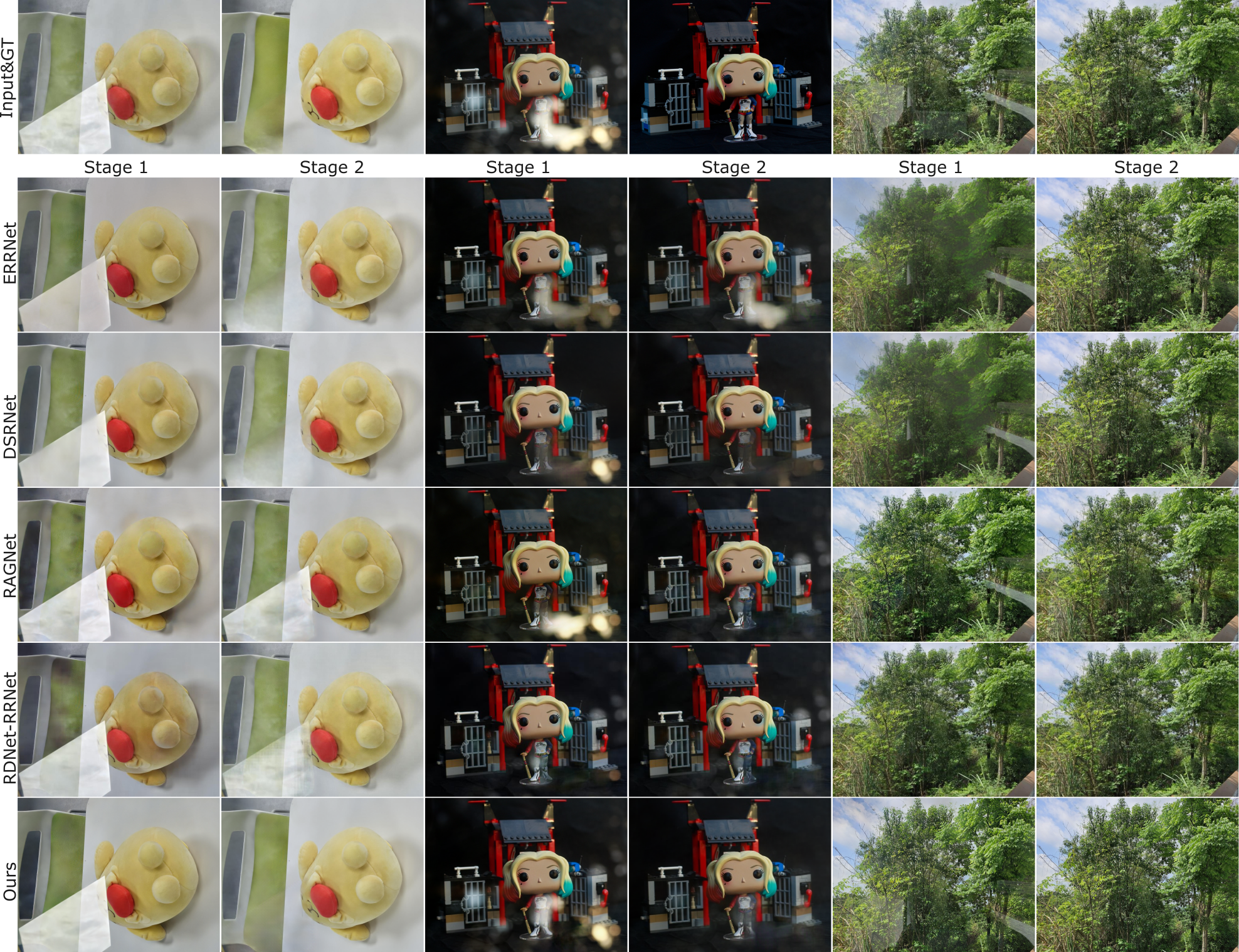}
    \caption{Comparison of visual results from ERRNet~\cite{wei2019single}, DSRNet~\cite{hu2023single}, RAGNet~\cite{li2023two}, and RDNet-RRNet~\cite{zhu2024revisiting}, and our NAFNet-based baseline across \textbf{S1} and \textbf{S2}~(before and after fine-tuning on OpenRR-1k dataset). }
    \vspace{-15pt}
    \label{fig:visualization}	 
\end{figure*}

\subsection{Experiment setting}
To conduct a comprehensive benchmark evaluation on the OpenRR-1k dataset, we proposed a new NAFNet-based baseline model by adapting the widely-used restoration architecture introduced in~\cite{chen2022simple}. For better representation learning ability, we expand the network’s bottleneck capacity by increasing the number of middle blocks from 1 to 12. Increasing the depth of the bottleneck allows for more sophisticated processing global information of image features, which improves the model’s ability to capture and handle complex reflection patterns. Then, we conducted two sets of evaluation experiments:

\textbf{S1:} We directly test the proposed baseline model and four state-of-the-art SIRR methods: ERRNet~\cite{wei2019single}, DSRNet~\cite{hu2023single}, RAGNet~\cite{li2023two}, and RDNet-RRNet~\cite{zhu2024revisiting} on the proposed OpenRR-1k$_{val}$ and OpenRR-1k$_{test}$ with the publicly available pre-trained weights, to verify the generalization ability of existing methods on challenging real-world data. We also provide the results of all reflection removal methods on three widely used reflection removal benchmarks: 20 test images from Real, 20 test images from Nature, all three sub-datasets from SIR$^2$ (a total of 454 test images).

\textbf{S2:} We fine-tuned the proposed baseline model and four comparison methods with the OpenRR-1k$_{train}$ dataset to verify whether the proposed dataset could improve the model generalization ability. 

The evaluation metrics we used are PSNR, SSIM, and LPIPS, where higher PSNR and SSIM values, and lower LPIPS values, indicate better performance.

\subsection{Implementation details}
The experiments were implemented using the PyTorch platform and were trained on a NVIDIA A100 GPU. For baseline model training in experiment \textbf{S1}, we trained the network for 60 epochs following~\cite{wei2019single}, employing the Adam optimizer. The initial learning rate was set to $10^{-4}$, which was halved at epoch 30 and further reduced to $10^{-5}$ at epoch 50. During the training, we used a combination of synthetic and real-world data. Specifically, synthetic data were generated from 7,643 images in the PASCAL VOC dataset~\cite{everingham2010pascal} following the image synthesis approach in~\cite{fan2017generic}. 289 training images in Real~\cite{zhang2018single} and Nature~\cite{li2020single} were included as real-world data. In experiment \textbf{S2}, we expanded the training set by incorporating training set of the proposed OpenRR-1k dataset. The model was then trained for an additional 100 epochs, starting with a learning rate of $5 \times 10^{-6}$, which was reduced to $2 \times 10^{-6}$ at epoch 50.

For the loss function, we used a weighted sum of pixel loss, feature loss, and adversarial loss~\cite{wei2019single, zhang2023benchmarking} as follows:
\vspace{-2pt}
{\small
\begin{align}
\mathcal{L}_{pixel} &= \alpha \left \| T^\star - T \right \| ^2_{2} \\ \nonumber
    &+\beta \left (  \left \| \nabla _{x}T^\star - \nabla _{x}T  \right \|_{1} + \left \| \nabla _{y}T^\star - \nabla _{y}T  \right \|_{1} \right )  \\
\mathcal{L}_{feat} &= \sum_{l}\lambda_{l}\left \| \Phi_{l}\left ( T \right ) - \Phi_{l} \left ( T^\star \right ) \right \| _{1}  \\ 
\mathcal{L}_{adv} &= -\log\left ( \sigma\left ( C\left ( T \right ) -C\left ( T^\star \right )  \right )  \right ) \\ \nonumber
&-\log\left ( 1- \sigma\left ( C\left ( T^\star \right ) -C\left ( T \right ) \right ) \right )  \\
\mathcal{L}_{all} &= \mathcal{L}_{pixel} + \lambda_{1}\mathcal{L}_{feat} + \lambda_{2}\mathcal{L}_{adv}
\end{align}
}\noindent
where $T$ represents ground truth while $T^\star$ denotes the estimated reflection-free transmitted image, $\alpha$ and $\beta$ are constants and set to 0.2 and 0.4, respectively. $\Phi_{l}$ represents layer $l$-th in the pre-trained VGG-19 network (specifically the layers ``conv2\_2", ``conv3\_2", ``conv4\_2", and ``conv5\_2"). $\lambda_{l}$ is the balancing weight for each layer; $\sigma\left ( \cdot \right )$ is the sigmoid function, $C\left ( \cdot \right )$ is the non-transformed discriminator function; and the weights $\lambda_{1}$ and $\lambda_{2}$ are set to 0.1 and 0.01 for $\mathcal{L}_{feat}$ and $\mathcal{L}_{adv}$, respectively. 

\subsection{Comparison with State-of-the-arts}
In this section, we compare our NAFNet-based baseline model with four state-of-the-art SIRR methods: ERRNet~\cite{wei2019single}, DSRNet~\cite{hu2023single}, RAGNet~\cite{li2023two}, and RDNet-RRNet~\cite{zhu2024revisiting} without and with introducing the proposed OpenRR-1k dataset for training (i.e., \textbf{S1} and \textbf{S2}). We summarize the results in Table~\ref{tab:2}. As shown in Table~\ref{tab:2}, under training strategy \textbf{S1}, DSRNet and RDNet-RRNet demonstrate particularly strong results, with DSRNet achieving the best performance on SIR$^2$ dataset (PSNR: 25.69, SSIM: 0.926, LPIPS: 0.088) and RDNet-RRNet performing well on Nature dataset (PSNR: 26.04, SSIM: 0.846, LPIPS: 0.100). Our baseline model performs well on both Nature dataset (PSNR: 25.44, SSIM: 0.844, LPIPS: 0.115) and Real dataset (PSNR: 23.25, SSIM: 0.824, LPIPS: 0.163). It is worth noting that for all methods (including our baseline), the PSNR and SSIM values are lower, and the LPIPS values are higher, compared to those of the unprocessed input image in the OpenRR-1k dataset. This indicates that despite achieving competitive performance on existing widely used benchmarks, current methods perform poorly when applied to ``true" real-world data. This supports our earlier analysis that existing datasets have significant limitations, specifically a lack of sufficient diversity and high-quality genuine data. This gap has become a major bottleneck in the field, which is our motivation for creating the OpenRR-1k dataset.

Furthermore, under training strategy \textbf{S2}, we observe a generally enhanced performance across all methods. Specifically, our baseline model achieves the best results on multiple datasets, including OpenRR-1k$_{val}$ (PSNR: 31.74, SSIM: 0.965, LPIPS: 0.040) and OpenRR-1k$_{test}$ (PSNR: 31.93, SSIM: 0.964, LPIPS: 0.038). DSRNet maintains strong performance, particularly showing a better result on Real (PSNR: 25.00, SSIM: 0.844, LPIPS: 0.131) and SIR$^2$ (PSNR: 25.83, SSIM: 0.928, LPIPS: 0.080). Moreover, when comparing the results between \textbf{S1} and \textbf{S2}, we can clearly observe that after fine-tuning on the diverse and high-quality genuine data, performance improvements are consistent across almost all methods and benchmarks. For instance, a significant gain is observed in our method's performance on the OpenRR-1k dataset, where PSNR increased by over 7 dB, SSIM increased by over 0.04, LPIPS improved by over 0.06 on both validation set and test set. This highlights the effectiveness of our OpenRR-1k dataset, meanwhile, demonstrates that the potential of existing models can be more fully realized when trained with our dataset.

Fig.~\ref{fig:visualization} further presents a comparison of the visual reflection removal results from competing methods and our baseline model across \textbf{S1} and \textbf{S2}, representing the results before and after fine-tuning on our OpenRR-1k dataset. In the first image (see the first/second column), all models show limitations in handling large areas of reflection before fine-tuning. The result of RDNet-RRNet even introduces color distortion, which impairs natural color reproduction of the toy surface. After fine-tuning, all methods significantly reduce reflections and avoid color distortion. As another example (see the third/fourth column), under challenging low-light conditions and multiple strong reflection sources, RDNet-RRNet removes the major strong reflections in \textbf{S1}, but leaves some weak residual reflections. In contrast, our baseline model in \textbf{S2} removes all reflections, resulting in a more convincing outcome. Regarding the last example (see the fifth/sixth column), although ERRNet and DSRNet lose high-frequency details in \textbf{S1}, all models successfully suppress reflections and preserve high-frequency transmission details in the results after fine-tuning.

%% file: 5-Conclusion.tex
\section{Conclusion}
\label{sec:conclusion}
In this paper, we propose a novel reflection removal pipeline that addresses a long-standing challenge in this field from a new perspective. Our pipeline provides researchers with a more convenient way to collect diverse, high-quality true real-world data samples at low cost. Using this pipeline, we constructed the OpenRR-1k dataset and conducted a comprehensive benchmark evaluation. Experimental results demonstrate the effectiveness and generalizability of our dataset in enhancing the performance of current state-of-the-art SIRR methods, unlocking their full potential. Future work will focus on creating a larger-scale dataset.